\documentclass[letterpaper]{article}
\usepackage{aaai}
\usepackage{times}
\usepackage{helvet}
\usepackage{courier}
\usepackage{color}
\usepackage{graphicx} 
\usepackage{booktabs} 
\usepackage[T1]{fontenc}
\usepackage{mathtools} 
\usepackage{amsmath}
\usepackage[normalem]{ulem}
\usepackage[backend=biber,style=authoryear]{biblatex}
\usepackage{wrapfig,lipsum,booktabs}
\usepackage{subcaption}
\usepackage{wrapfig}

\definecolor{orange}{rgb}{1,0.5,0}

\begin{document}
%
\title{Reducing the dilution: An analysis of the information sensitiveness of capsule network with a practical improvement method}
\author{
  Yang, Zonglin\\
  \texttt{zlyang@hust.edu.cn}
  \And
  Wang, Xinggang\\
  \texttt{xgwang@hust.edu.cn}
} 

\maketitle
\begin{abstract}
\begin{quote} 
Capsule network has shown various advantages over convolutional neural network (CNN). It keeps more precise spatial information than CNN and uses equivariance instead of invariance during inference and highly potential to be a new effective tool for visual tasks. However, the current capsule networks have incompatible performance with CNN when facing datasets with background and complex target objects and are lacking in universal and efficient regularization method. 

We analyze a main reason of the incompatible performance as the conflict between information sensitiveness of capsule network and unreasonably higher activation value distribution of capsules in primary capsule layer. Correspondingly, we propose a practical improvement method by restraining the activation value of capsules in primary capsule layer to suppress non-informative capsules and highlight discriminative capsules. In the experiments, the method has achieved better performances on various mainstream datasets. In addition, the proposed improvement methods can be seen as a suitable, simple and efficient regularization method that can be generally used in capsule network. 
\end{quote}
\end{abstract}

\section{1. Introduction}
Capsule network\autocite{sabour2017dynamic} is a novel structure for visual tasks. It has shown huge potential among multiple data sets and it is drawing more and more attention. Capsule network has various distinctive advantages that convolutional neural network(CNN) doesn't possess, like it replaces max-pooling, a method  commonly used in CNN that can result in spatial information loss, so that capsule network can keep the spatial relationship and therefore can make more reasonable classification result. In addition, capsule network has the potential to resist white box adversarial attack\autocite{hinton2018matrix} and can extract structured features that have equivariance characteristic rather than invariance which may lead to many good properties \autocite{kondor2018covariant} \autocite{cohen2016group} \autocite{kondor2018generalization}. 

However, multiple obstacles are on the road of capsule network, preventing it from being largely implemented and disabling researchers to utilize its advantages.  One main challenge is that current implementation\autocite{sabour2017dynamic} of the general idea of capsule net\autocite{hinton2011transforming} has shown high susceptibility to background information\autocite{sabour2017dynamic}. We attribute one main reason to this problem as the conflict between  information sensitiveness of capsule network(will be discussed in Section 5.1) and unreasonable activation value distribution of capsules in primary capsule layer. Specifically, capsule network will pass each capsule's information to next layers in a full magnitude of its activation value(will be further discussed in Section 5.1) and lacks suitable mechanism of selecting discriminative information from outputs of each layer(CNN has ReLU\autocite{glorot2011deep} as a comparison). But original squash function encourages the capsules with low energies to have large activation values, resulting in an unreasonable higher distribution of activation values and thus many capsules encode irrelevant information receive larger activation values than they should (will be further discussed in Section 3) and finally causes information entangling and irrelevant information disturbing (especially when facing images with background information) and therefore leads to a relatively poor performance.

The improvement methods we propose is more selective to assign high activation value and generally assigns lower activation values to pick and attach more importance to the more discriminative capsules to improve the performance of capsule network. However it may seem difficult to implement, instead of developing a complicated computational block that may cause capsule net to consume a lot larger amount of computation, we have developed them in quite a simple way. In our improvement methods, capsules that correspond to an irrelevant spatial position(like background) or calculated from not matched convolution weights) that usually have small activation value will be suppressed and the network will be able to learn a more discriminative and disentangled representation(will be further discussed in Section 5.2). With our methods, we have observed a steady performance gain in multiple mainstream datasets.

In addition, our method can be seen as a suitable regularization method for capsule network since that our method represses the strength of connections of capsules in adjacent capsule layers and have the effect of forcing the net to learn a more sparse and more discriminative representation(will be discussed in Section 5.2). In addition, our experiments show that common regularization methods as dropout and weight decay both have a negative effect on the performance of capsule network(shown in Section 4.2). Furthermore, reconstruction as a regularization method has limitations only where can be helpful in relatively simple input data like MNIST. The limitations of all previous regularization methods may make our method the only regularization method that is suitable and applicable for capsule network. In addition, also considering the simplicity and efficiency of our methods, we argue that our method can be used as a common regularization method for capsule network. 

\section{2. Capsule Network}
\subsection{2.1 Main Characteristics}
The current concrete implementation of capsule network\autocite{sabour2017dynamic} (also called capsnet in the following sections) can be seen as a partial CNN model combined with an implementation of general capsule idea. Capsnet mainly has two innovations compared with CNN. The first is that capsnet no longer uses pooling operation (using Routing instead) so that capsnet can maintain the spatial relations between object parts in input image. The other innovation is that capsnet uses feature vector to represent an object instead of feature scalar. This change enables capsnet to have a representation that is more to equivariance instead of invariance, mainly have internal changes in capsule that is highly related to changes of the object in the input image. For instance, when object in image rotates or switches to a different color, the internal relations between scalar components in a capsule (by capsule, we refer to an activation vector or in other words a feature vector) that represents this entity basically holds unchanged, but the whole capsule vector may rotate or translates into a corresponding scale.

\subsection{2.2 Main Structure}
Capsnet firstly uses convolution of $9\times 9$ convolutional kernel on the input image to get the first layer of feature map (first convolutional layer), then uses another convolution of $9\times 9$ convolution kernel to get the second layer of the feature map. Capsnet then slices the second layer of feature map to get a number of feature vectors(or feature matrices) called capsules and assigns these capsules each with an activation value. Specifically, in \autocite{sabour2017dynamic}, these activation values are assigned with the output of squash function(Eqn.~\eqref{eq:squash_detailed}) with $\|\textbf{s}_j\|$(explained in section 2.3) as the length of sliced feature vector. 

These capsules compose of the first capsule layer called primary capsule layer. Given capsules in primary capsule layer, capsules in its next layer is generated via a routing mechanism. In \autocite{sabour2017dynamic}, this next capsule layer is called digit capsule layer, which is the last layer of capsnet, and will be used to make a classification decision using the activation value of each of its capsules, each of which corresponds to a different class. 

\subsection{2.3 Dynamic Routing and Votes} 
Dynamic Routing is an implementation of the routing mechanism that is used to calculate the values of capsules in next capsule layer given the value of capsules in former capsule layer. Generally, each capsule keeps one activation vector and each capsule in former capsule layer will generate an activation vector proposal (will be called vote in the following sections) for every capsule in next capsule layer. Each capsule in next capsule layer collects all votes to it and use them to calculate a cluster centroid, which will be further processed to be used as the activation vector of this capsule in next capsule layer. 

To describe its mechanism more specifically, we can first define some symbols. We use $\textbf{u}_i$ as activation vector kept by capsule $i$ in one capsule layer, $\textbf{u}_j$ as activation vector kept by capsule $j$ in the next capsule layer. $\textbf{v}_{j|i}$ is the vote capsule $i$ generates for capsule $j$, generated by the product of $\textbf{u}_i$ and a learnable weight matrix $\textbf{w}_{j|i}$. It can be shown in Eqn.~\eqref{eq:vote}.

\begin{equation}
\label{eq:vote}
\textbf{v}_{j|i} = \textbf{w}_{j|i} * \textbf{u}_i
\end{equation}

$\textbf{c}_{j|i}$ is the weight for $\textbf{v}_{j|i}$ to get a weighted sum of a summary proposal $\textbf{s}_j$ for $\textbf{u}_j$, gotten from a softmax of $b_{j|i}$. $b_{j|i}$ are all initialized to $0$ at the beginning. Previous process can be expressed in Eqn.~\eqref{eq:ini_b}, Eqn.~\eqref{eq:softmax} and Eqn.~\eqref{eq:sum_for_vote}:

\begin{equation}
\label{eq:ini_b}
b_{j|i} \longleftarrow 0
\end{equation}

\begin{equation}
\label{eq:softmax}
c_{j|i} = softmax(b_{j|i}, \text{axis}=j) 
\end{equation}

\begin{equation}
\label{eq:sum_for_vote}
\textbf{s}_j = \sum_{i}(c_{j|i} *\textbf{v}_{j|i})
\end{equation}

Squash function is operated on the weighted sum of votes from former layers($\textbf{s}_j$) to make the range of each capsule vector to be $[0,1)$, which can thus be expected to represent the probability that the entity represented by the capsule is present in the current input. It can be represented by Eqn.~\eqref{eq:squash_detailed}:
\begin{equation}
\label{eq:squash_detailed}
\textbf{u}_j = squash(\textbf{s}_j) = \frac{\|\textbf{s}_j\|^2}{1+\|\textbf{s}_j\|^2} \frac{\textbf{s}_j}{\|\textbf{s}_j\|}
\end{equation}

Now we have got $\textbf{u}_j$, but it is only a result after one round of routing method. \autocite{sabour2017dynamic} has shown that three round of routing can lead to the best performance. To do another round, we don't need to change $\textbf{u}_i$ or $\textbf{v}_{j|i}$ but to compute a more suitable $\textbf{c}_{j|i}$ by updating $\textbf{b}_{j|i}$. It can be represented by Eqn.~\eqref{eq:update_b}.
\begin{equation}
\label{eq:update_b}
b_{j|i} = b_{j|i} + \textbf{v}_{j|i}*\textbf{u}_{j}
\end{equation}
Then we can go on routing from Eqn.~\eqref{eq:softmax} to Eqn.~\eqref{eq:squash_detailed} to get another round of routing to update activation vectors for capsule $j$. 

\vspace{-2mm}
\section{3. Information Sensitiveness of Capsule Network}
\subsection{3.1 Analysis}
\subsubsection{Connections between adjacent capsule layers}
As that we can only calculate $c_{j|i}$ during the inference process and therefore it is not fixed and learnable, connections between capsules in adjacent capsule layers are not fixed and are variant depending on different input data.

However, we can express the strength of a connection between capsule $i$ and capsule $j$, also we can call it the importance, or a fixed weight coefficient as neural network has between capsule $i$ for capsule $j$ for a fixed input, as the norm of the product of $c_{j|i}$ and $\textbf{v}_{j|i}$, as shown in Eqn.~\eqref{eq:connection_i_j}.

\begin{equation}
\label{eq:connection_i_j}
\|connection_{ij}\| = \|c_{j|i}\|*\|\textbf{v}_{j|i}\|
\end{equation}

\subsubsection{Quantitative analysis of sensitiveness}
The mechanism of routing mechanism has been shown in Section 2.2. Our attention mainly focuses on how weights are calculated. What we want to point out is that softmax, which is used to calculate the distribution weights of capsule, is operated on all weights of vote generated by capsule from the former capsule layer. Therefore in this process, every capsule in the former layer spread itself in full magnitude, since the sum of the output of softmax is $1$, of its activation value to the next capsule layer. We argue that this phenomenon can result in the sensitiveness of capsule network.

Specifically, we can understand this sensitiveness through how one capsule can influence capsules in the next capsule layer. We have given an expression of the strength of the connection between capsules in adjacent capsule layers in Section 2.3. We can approximately consider the influence of a capsule to a capsule in the next capsule layers as this strength of the connection, just as that the influence of a neuron to a specific neuron in the next layer can be measured as the norm of their connection in multi-layer perception, which is usually a fixed weight. Therefore the influence between capsule $i$ and capsule $j$ (recall that capsule $j$ represents a capsule in the next capsule layer of capsule $i$) can be approximated as Eqn.~\eqref{eq:influenceij_approx}.

\begin{equation}
\label{eq:influenceij_approx}
\|Influence_{ij}\| \approx \|connection_{ij}\| = \|c_{j|i}\|*\|\textbf{v}_{j|i}\|
\end{equation}

Also as Eqn.~\eqref{eq:vote}, we can get Eqn.~\eqref{eq:influenceij_unraveled}.

\begin{equation}
\label{eq:influenceij_unraveled}
\|Influence_{ij}\| \approx \|c_{j|i}\|*\|\textbf{w}_{j|i}\| *\|\textbf{u}_{i}\|
\end{equation}

We can express the total influence of capsule $i$ as the sum of $\text{influence}_{ij}$, shown as Eqn.~\eqref{eq:influence_i}.

\begin{equation}
\label{eq:influence_i}
\|Influence_{i}\| = \sum\limits_{j}{\|Influence_{ij}\|}
\end{equation}

In addition, since Eqn.~\eqref{eq:sum_c} 
\begin{equation}
\label{eq:sum_c}
\sum\limits_{j} \|c_{j|i}\| = 1
\end{equation}
and that our experiment shows that for a fixed $i$, $\|\textbf{w}_{j|i}\|$ for all $j$ usually have similar norm, which we can denote as $\|\textbf{w}_i\|$, we can get Eqn.~\eqref{eq:influence_sum};

\begin{equation}
\label{eq:influence_sum}
\|Influence_i\| \approx \sum\limits_{j}{\|c_{j|i}\|*\|\textbf{v}_{j|i}\|} = \|\textbf{w}_i\|*\|\textbf{u}_i\|
\end{equation}

In addition, for a trained capsule network, the value of transformation matrix $\textbf{w}$ is fixed, thus Eqn.~\eqref{eq:constant_w} establishes.

\begin{equation}
\label{eq:constant_w}
\|Influence_i\| \approx constant*\|\textbf{u}_i\|
\end{equation}

In conclusion, with the current routing mechanism, we can approximately consider the influence of a capsule to the next layer as the product between a constant and its activation value. If capsules that encode irrelevant information, no matter because of its spatial position or its weights of convolution, cannot be suppressed to a reasonably small value, also considering that most of capsules can be not important for a specific input image as shown in Fig~\ref{fig:coeff}, then these many relatively indiscriminative capsules can have a big integral negative influence on the network. Specifically, it may cause a large proportion of chaotic information passing to the next capsule layer and finally leads to a bad result.

\subsubsection{Analysis}
We have two points to articulate based on the conclusion we have got (Eqn.~\eqref{eq:constant_w}), which can give us a direct understanding of information sensitiveness of capsule network and its vulnerability to capsules that encodes irrelevant information and possesses large activation value. 

One is that on contrary to our intuition, routing mechanism doesn't give votes that are far away from its cluster centroid very small weights such that irrelevant information won't have a big impact on it, instead, for each particular capsule, no matter whether information it encodes are relevant or helpful for a visual task, it will have an influence that is in proportion to its activation value to next layers of capsule network. 

The other is that although CNN can also be seen as a similar process, which is that a neuron will pass its information to neurons in the next layer, CNN has sparsity method like ReLU that can potentially filter out irrelevant information from a lower layer, while capsule network doesn't have it. Specifically, inference of CNN can be seen as the process of information being passed, purified and explained -- Background features are continuously being filtered out as convolution can be interpreted as a template matching process\autocite{yuille2018deep} and information like background information that does not match is highly possible to be calculated to a negative value and will be set to zero when passing ReLU\autocite{glorot2011deep} layers. So that information of target object will be steadily accumulated and gradual explained from pixel level to semantic level. Finally, fully connected layers can, therefore, make a good decision based on these more-relevant-to-target-object semantic features. 

The lack of sparsity can cause a result we call as information sensitiveness of capsule network. That is to say, information kept in capsule will be passed through the following net with a full magnitude of its activation value, no filtered out as CNN does and no matter whether this information is helpful or irrelevant for the final layer to make a decision. This no-filtering-out irrelevant information may be entangled with the effective information and make capsnet more and more confused as information passing through the network and as a result, the final layer of capsnet can't make a good decision from the information it receives.

\vspace{-3mm}
\subsection{3.2 More Proper Regularization}
\autocite{sabour2017dynamic} uses reconstruction as the only regularization method uses in capsule network. However, using reconstruction as a regularization method has multiple shortcomings. Firstly, it can only be useful for simple input data like MNIST. If we use it in relatively complex data like CIFAR10, the reconstruction image will be very blurry and reconstruction is not helpful to get better performance. Secondly, the reconstruction method adds much additional computation and will make capsule network that is originally quite slow more consuming to train and inference. 

Besides image reconstruction, the most commonly used regularization method in computer vision models are weight decay~\autocite{krogh1992simple} and dropout~\autocite{hinton2012improving}. However, in the current capsule network, weight decay and dropout are both shown to have a negative effect on performance, as has shown in Section 4.2. For weight decay, we conjecture that weight for generating votes do not represent more as a connection but more as a complicated transformation. Therefore instead of the effect of deemphasizing every single connection and achieving better generalization, weight decay prevents weights from learning a suitably complex enough transformation from capsules to votes, greatly decreases the capacity of capsule net and therefore decreases the performance of capsule net. For dropout, a potential explanation is that dropping some of capsules do not represent more on dropping a set of features in semantic level so that network with dropout is compelled to learn more independent features, but more on dropping feature vectors in spatial information level, since that different capsules correspond to different image areas and primary capsule is in shallow layer that cannot encode features semantic-level enough. Therefore dropout is kinds of like randomly erase the information of a random region of an image. The consequence is that by using dropout, instead of finding other semantic level feature to help to make a decision, capsules are compelled to learn features around them (e.g. features in the margin of its reception field) in case the surrounded features being dropped out. This will cause denser connections between capsules in adjacent capsule layers and may lead the capsules to be learned more dependently on each other, which is completely contrary to the idea of dropout and regularization.

\begin{figure}[!htp]
\centering   
\includegraphics[width=1\linewidth]{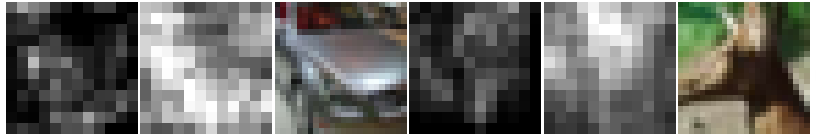}   
\caption{Comparison activation map between with PA and without PA (left: with PA; middle: original squash without PA; right: original picture)}
\label{fig:Comparison_in_activation_map_between_PA_and_nonPA}   
\end{figure}  

\begin{figure}[h]
\centering   
\includegraphics[width=1\linewidth]{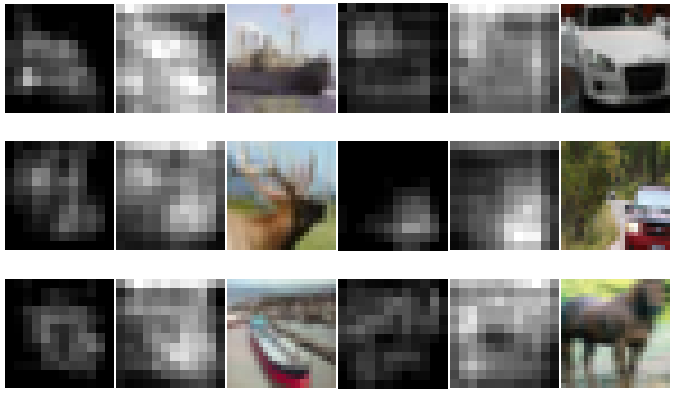}   
\caption{Comparison activation map between CI-squash and original squash (left: CI-squash; middle: original squash ; right: original picture)}
\label{fig:Comparison_activation_map_between_CI-squash_and_original_squash}   
\end{figure}

However, our two improvement methods, PA and CI-squash, can both be seen as regularization methods since they are both compatible with the idea of regularization that can give a suppression effect to connections of capsules in adjacent layers and lead to a sparser and more representative representation. Specifically, as Eqn.~\ref{eq:influenceij_approx} shows, we suppress the activation values, therefore we can indirectly suppress the connection strength between capsules in adjacent capsule layers. In addition, our experiment results shown in Fig.~\ref{fig:Comparison_in_activation_map_between_PA_and_nonPA} and Fig.~\ref{fig:Comparison_activation_map_between_CI-squash_and_original_squash} are also supportive to our hypothesis of being seen as a regularization method(leads to more sparse and more representative representation). In sum, we argue that our methods used in capsule network are actually more proper regularization methods for capsule network.

\section{4. Improvement methods for Capsule Network}
\begin{figure}[h]
\centering 
\includegraphics[width=0.88\linewidth]{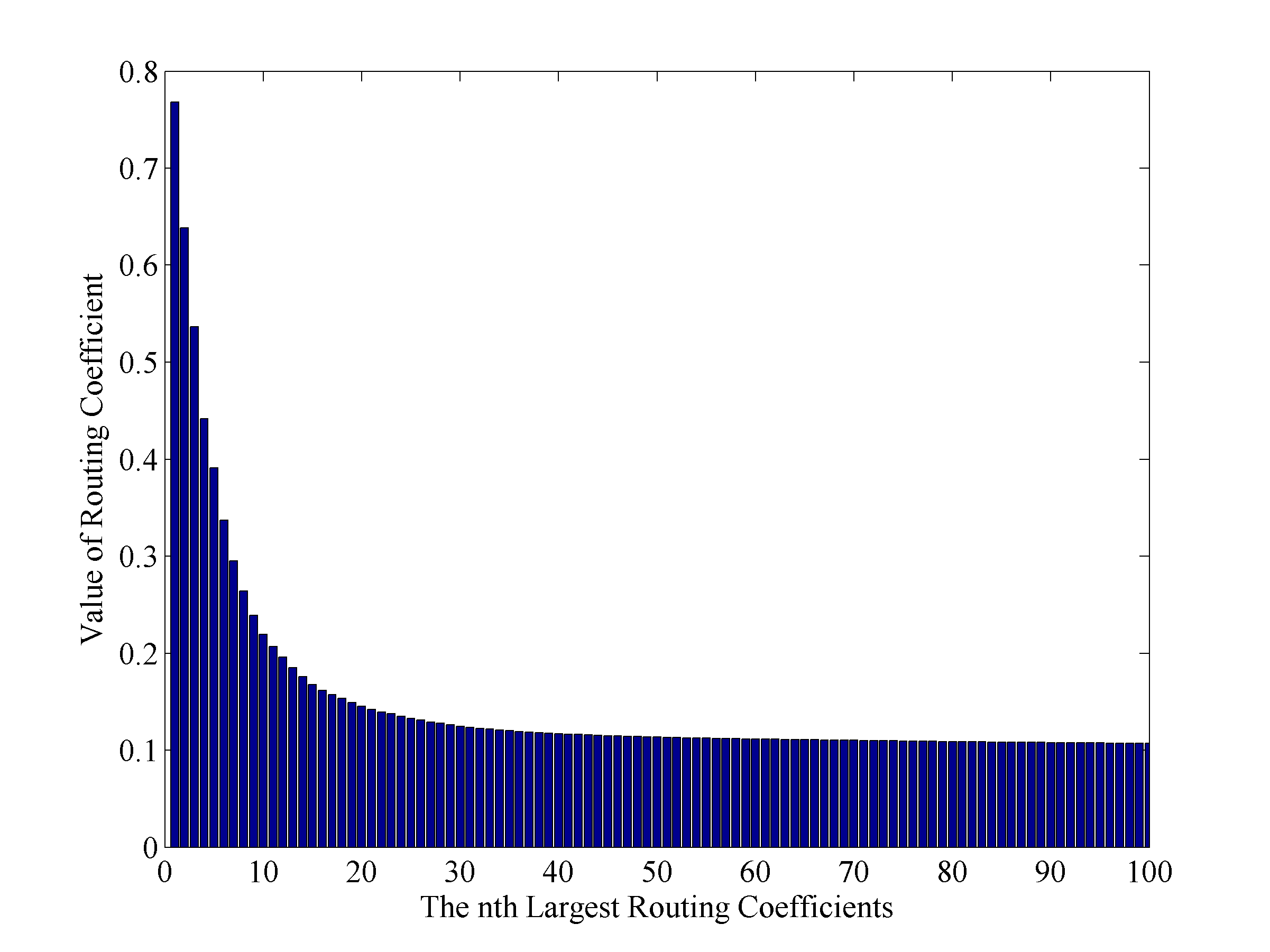}   
\caption{We get this figure by firstly retrieving the biggest routing coefficient of each capsule, then put them in order(8*12*12 in total), mean these ordered series over test data set, and show the largest 100 of them.}
\label{fig:coeff}
\end{figure}

An important hypothesis of capsule network\autocite{sabour2017dynamic} is that the activation value of a capsule represents the probability that a specific type of entity such as an object or an object part exists. With this hypothesis, we can infer that there should be a large number of capsules that are not helpful to discriminate one specific object in an input image, given that different target objects have very different object parts. Our experiment shows the consistent result with our conjecture that when tested with CIFAR10 dataset. Specifically, only $20$ of $8*12*12$ primary capsules on average have largest weight coefficient $\max(c_{j|i})$ of $j$ with value that is more than $0.15$, as shown in Fig.~\ref{fig:coeff}. Note that if a capsule can't help to distinguish the $10$ classes, it should vote for capsules in digit capsule layer randomly in a weight of $0.10$. Therefore there could be a large amount of capsules that encodes information that is not helpful to get a classification result.

However, the original squash function used to generate activation value is prone to generate a high activation value even for a capsule that only has small $\|\textbf{s}_j\|$ as that the original squash function grows quickly in the beginning. In addition, considering that the current routing mechanism will pass information encoded by a capsule fully in its activation value to capsules in next layers which will not be filtered out even if that information is not helpful to make a classification decision (sensitiveness of routing mechanism, as will be further discussed in Section 5.1), it is reasonable to introduce sparsity to restrain capsules from easily getting a high activation value to prevent information disturbing and highlight capsules that are helpful to distinguish targets. Our two methods are both designed on this idea.

In this section, we will introduce two new methods that can introduce sparsity. We name them as Cubic-Increasing Squash(CI-squash) and Powered Activation(PA). They are all modified versions of original squash function(Eqn.~\eqref{eq:squash_detailed}). Squash function has been used in two circumstances, one is used in primary capsule layer to calculate the initial activation value of each capsule in primary capsule layer, the other is used in routing mechanism. However, for now, we only modify squash function in the former circumstance. Specifically, we only change the way how initial activation values are calculated for capsules in primary capsule layer. We show that with this small change we can achieve a significant performance gain for capsule network. 

\begin{figure}[!htp]
\centering   
\includegraphics[width=0.88\linewidth]{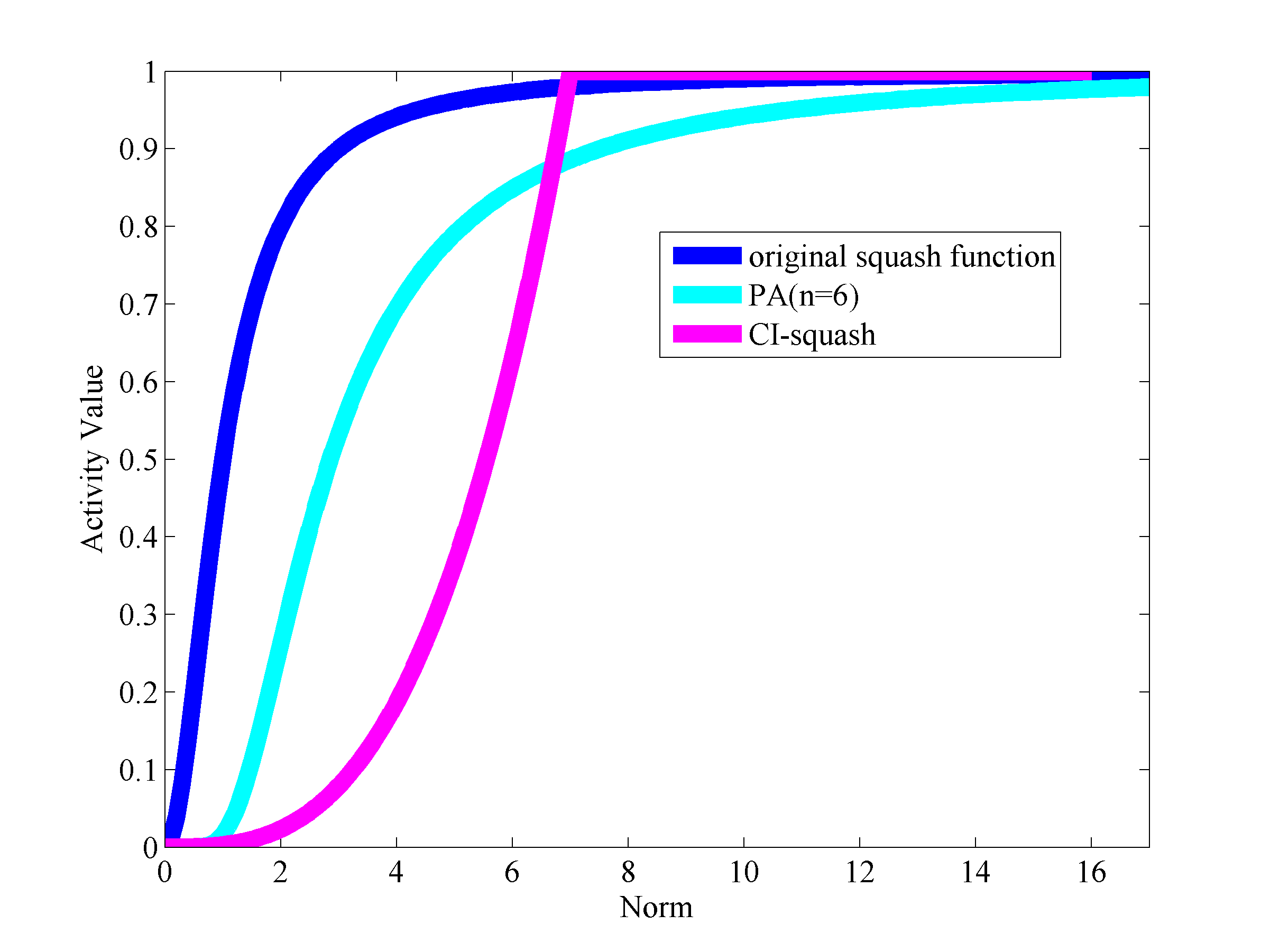}   
\caption{Comparison between original squash function and PA and CI-squash}
\label{fig:squash_PA_CIsquash}   
\end{figure}

PA and CI-squash, as both improvement methods, are both the important implementation of idea of trying to solve the information sensitiveness problem. Both methods can lead to a performance gain and they can excel each other in a different dataset or with different hyperparameter setting. Therefore we will introduce both methods in detail in this section, we will introduce how to choose these two methods in Section 4.3. Curves of original squash, CI-squash and PA are shown in Fig.~\ref{fig:squash_PA_CIsquash}.

\subsection{4.1 Cubic Increasing squash function (CI-squash)}
We design CI-squash as a new squash function who grows at the speed of cube at first so that activation vectors with small length will be assigned with very small values. When the input length of capsules is greater than its threshold(called bar in the following passage), the activation value will be set to $1$, as the input length is large enough that we can believe that there is an entity existed in that capsule. Its equation is shown in Eqn.~\eqref{eq:CI-squash}.
\begin{equation}
\textbf{u}_j = \{\frac{-\text{ReLU}(-\textbf{s}_j+bar)+bar}{bar}\}^3*\frac{\textbf{s}_j}{\|\textbf{s}_j\|}
\label{eq:CI-squash}
\end{equation}
\vspace{-1mm}
\subsection{4.2 Powered Activation (PA)}
Instead of using a new squash function, we try to figure out which element of new squash function does matter for capsule network. With that element, we can directly add it to the original squash function and we should observe a similar positive effect. 

PA can be seen as that key component. The procedure of PA can be explained as equation Eqn.~\eqref{eq:squash}, Eqn.~\eqref{eq:powern} and Eqn.~\eqref{eq:using_powern}. Original capsule network uses $\textbf{u}_j$ as activation vectors to pass to next computational block of routing mechanism. Our method is that we powered $\textbf{u}_j$ and instead we pass the output of it, $\hat{\textbf{u}}_j$ in specific, to next block. 
\begin{equation}
\textbf{u}_j = {squash}(\textbf{s}_j)
\label{eq:squash}
\end{equation}
\begin{equation}
Power_n(\textbf{x}) = \|\textbf{x}\|^n \frac{\textbf{x}}{\|\textbf{x}\|}
\label{eq:powern}
\end{equation}
\begin{equation}
\hat{\textbf{u}}_j = {Power_n}(\textbf{u}_j)
\label{eq:using_powern}
\end{equation}

\begin{figure}[h]
\centering   
\includegraphics[width=0.88\linewidth]{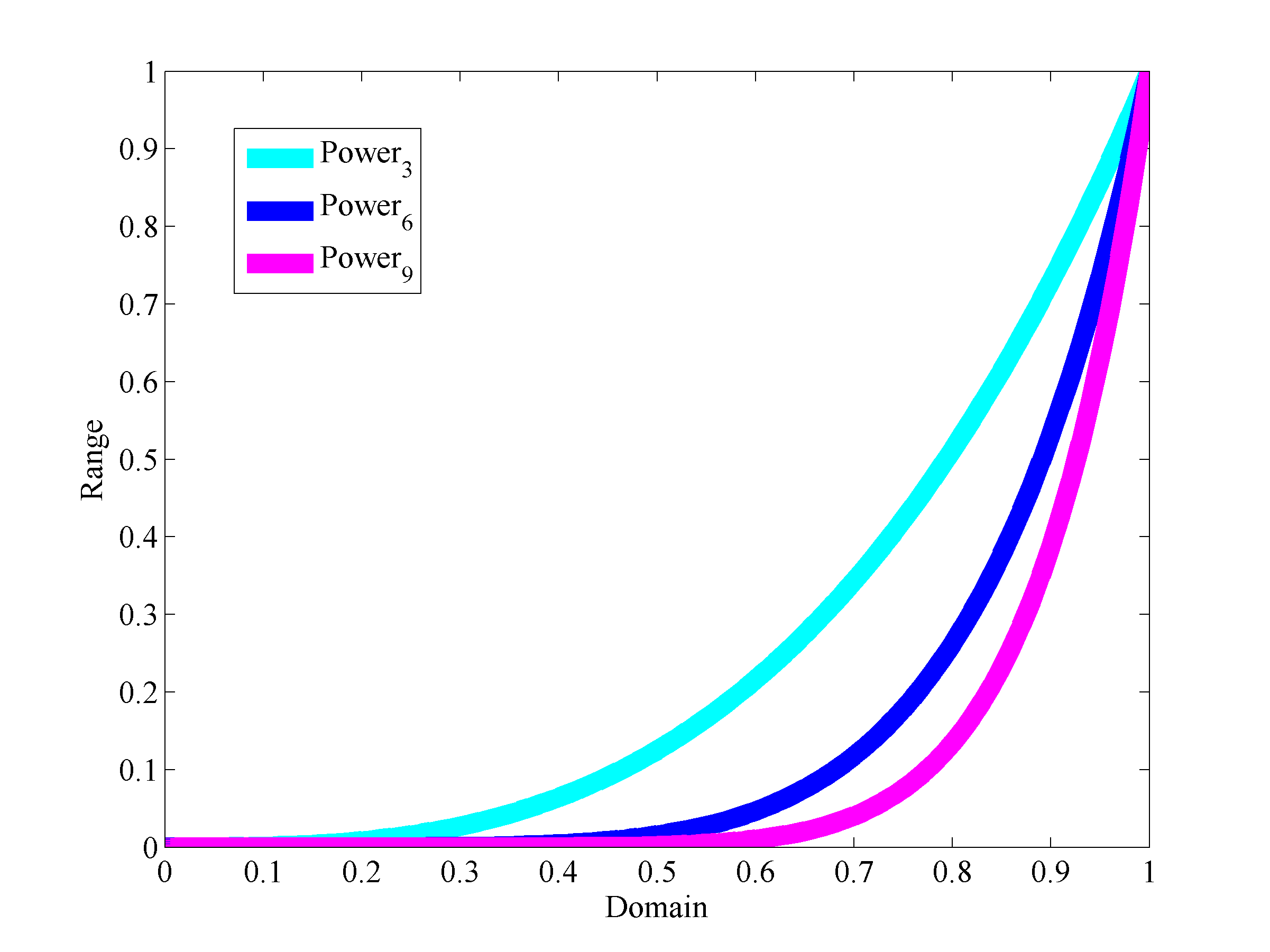}   
\caption{$Power_n(\cdot)$ functions}
\label{fig:powered_n_function}   
\end{figure}

The range of $\textbf{u}_i$ is $[0,1)$, thus $Power_n(\cdot)$ function in effective domain can be shown in Fig.~\ref{fig:powered_n_function}. We can see that it has much more suppression effect on small activation values than on large values. We expect $Power_n(\cdot)$ function can suppress capsules with small activation values while keep those with large activation values and to result in a sparsified activation values.

\section{5. Experiments}

This section we carry out experiments to validate various implementations of the sparsity capsule networks on multiple datasets, including MNIST \autocite{lecun1998gradient}, CIFAR10 \autocite{krizhevsky2009learning} and MultiMNIST \autocite{sabour2017dynamic}. To make fair comparisons with the original capsule network \autocite{sabour2017dynamic}, we use exactly the same setting with \cite{sabour2017dynamic}, including training/testing splitting, data augmentation, and hyper-parameter setting. In addition, to avoid other factors that may influence this comparison, such as difference of hardware or difference of accuracy calculating method, we also carry out experiments of original capsule network with the source code released by \autocite{sabour2017dynamic} on our own. Specifically as for accuracy calculating method, each result in this paper is obtained by averaging up around 40 checkpoints (1,500 steps gap between adjacent checkpoints and batch size is 128) generated after the model is converged(also not overfitting).

\subsection{5.1 Results on various dataset}

\subsubsection{MNIST and MultiMNIST}

\begin{table}[h]
\centering   
\caption{Error rate of experiments on MNIST and MultiMNIST}  
{
\begin{tabular}{c|cc}
\toprule
Method & MNIST & MultiMNIST\\
\midrule
capsnet+original squash & 0.3802\% & 5.82\% \\
\midrule
capsnet+CI-squash & 0.3604\% & 5.33\% \\
\midrule
capsnet+PA($n$=6) & \textbf{0.3308}\% & \textbf{5.12}\% \\ 
\bottomrule
\end{tabular}
}
\label{tab:mnist}
\end{table}

The results on MNIST are shown in Tab.~\ref{tab:mnist}. The results show that the capsnets with our improvement methods (capsnet+CI-squash and capsnet+PA($n$=6)) both outperform the original capsnet. 

For MultiMNIST, we follow the setting of \autocite{sabour2017dynamic}, generating 60M training images. But since limited resources, we only train it for 1.1 epochs and test it on 200K testing images(each MNIST image generate 20 MultiMNIST testing images). Although it is not as fully trained as \autocite{sabour2017dynamic} does, we achieve the state-of-the-art error rate of 5.12\%, better than the result reported by \autocite{sabour2017dynamic} as 5.20\%.

\subsubsection{CIFAR10}

CIFAR10 is more challenging than MNISTs for capsule networks since it contains both background information and more complex objects. In the experiments of CIFAR10, we compare the capsule networks with a different number of primary capsules. Results are shown in Tab.~\ref{tab:cifar}.

\begin{table}[h]
\centering   
\caption{Error rate of experiments on CIFAR10}  
{
\begin{tabular}{c|cc}
\toprule
Method &capsnet(primCaps8)& capsnet(primCaps64) \\
\midrule
 original squash&16.15\%&14.24\%  \\
\midrule
CI-squash&14.67\%&\textbf{13.72}\%  \\ 
\midrule
 PA($n$=6)&\textbf{14.31}\%&14.11\% \\
\bottomrule
\end{tabular} 
}
\label{tab:cifar}
\end{table}

\begin{figure*}[t!]
\centering
  \begin{subfigure}[t]{0.3\textwidth}
    \includegraphics[width=1\textwidth]{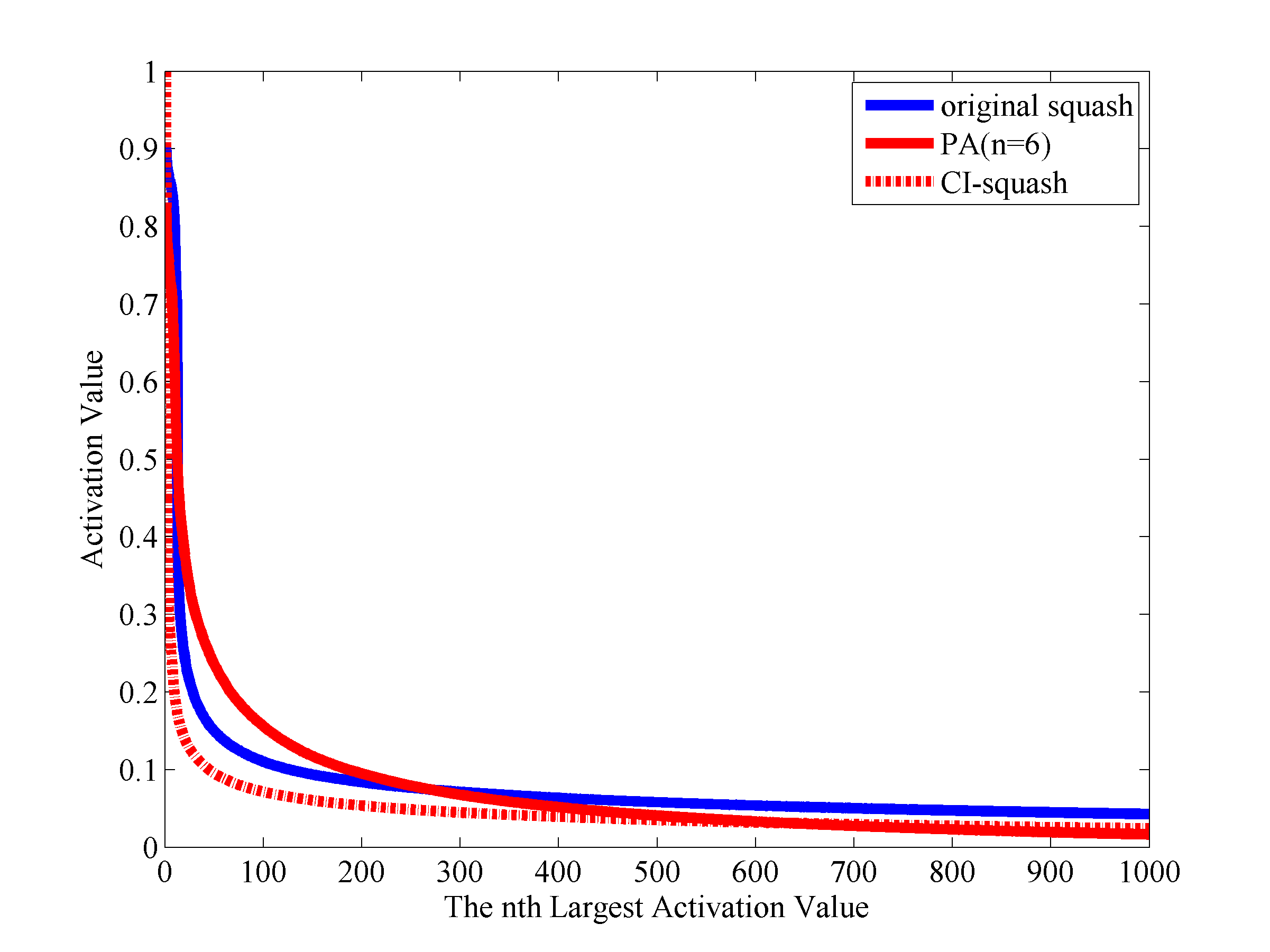}
     \caption{}
  \end{subfigure}
  \begin{subfigure}[t]{0.3\textwidth}
      \includegraphics[width=1\textwidth]{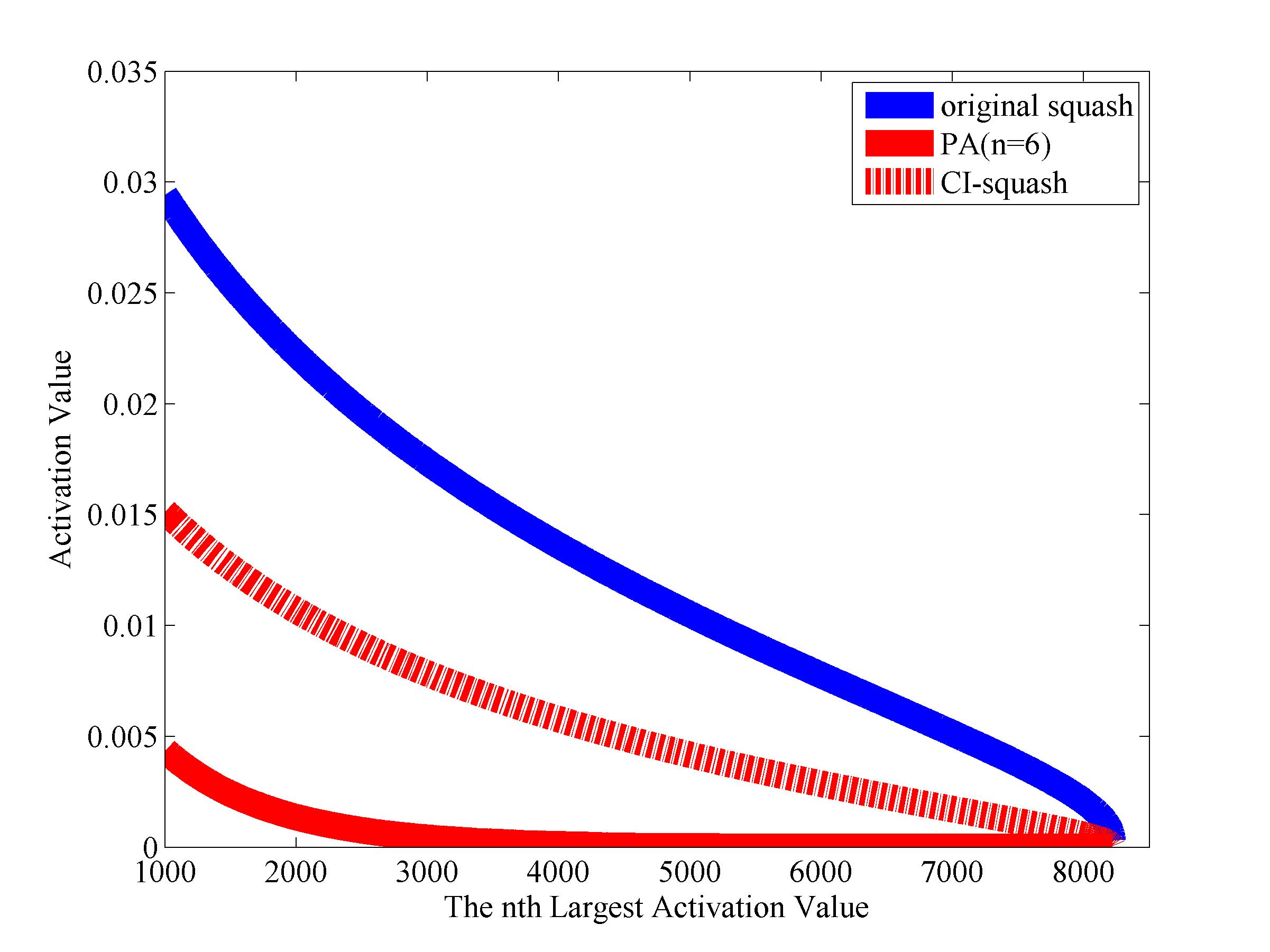}
    \caption{}
  \end{subfigure}
  \begin{subfigure}[t]{0.3\textwidth}
      \includegraphics[width=1\textwidth]{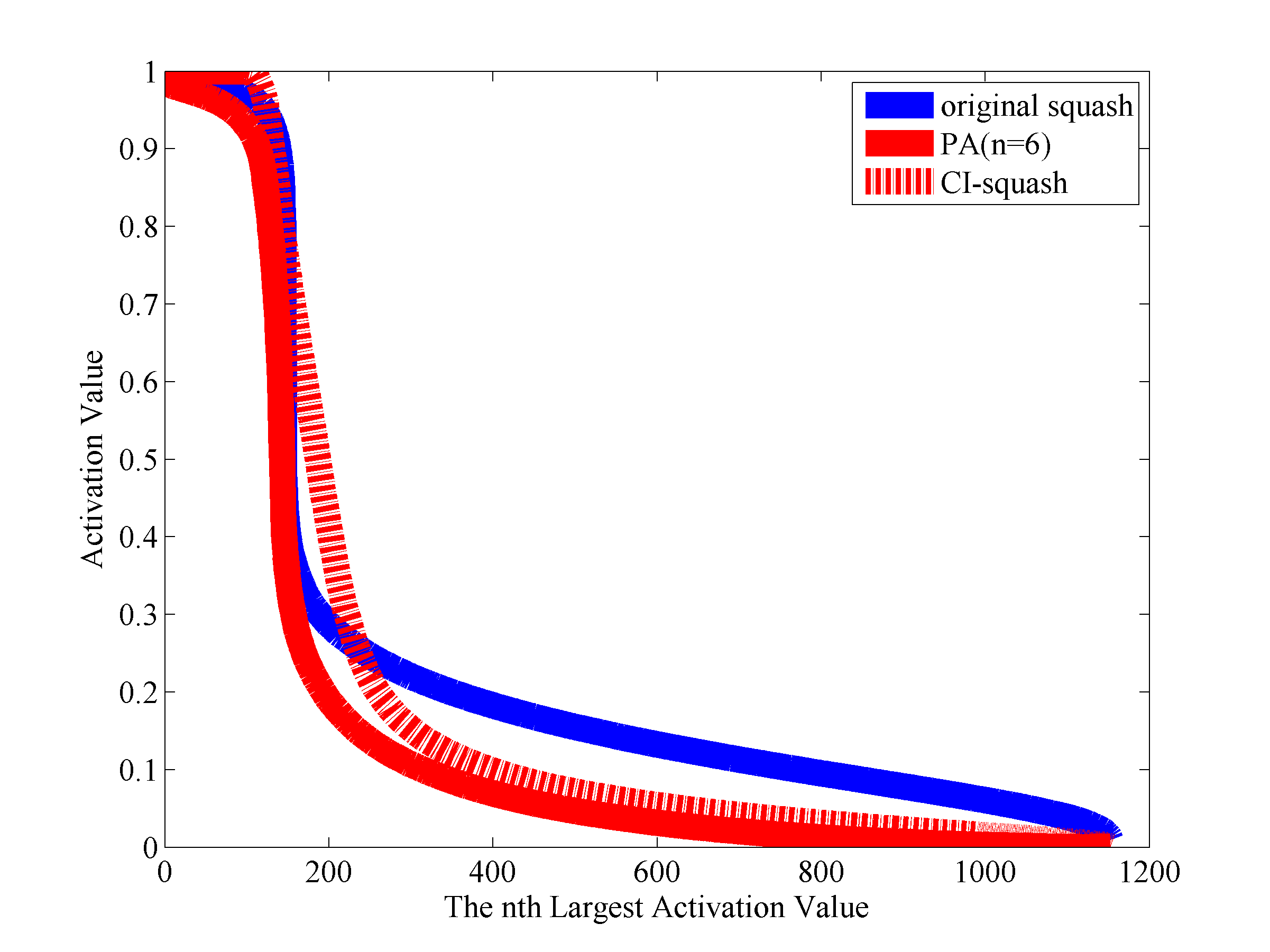}
    \caption{}
  \end{subfigure}
  \caption{(a): Comparison of the largest 116 of ordered activation value between original squash, CI-squash and PA with 64 primary capsule layers. (b): Comparison of full ordered activation value between original squash, CI-squash and PA with 64 primary capsule layers. (c): Comparison of full ordered activation value between original squash, CI-squash and PA with 8 primary capsule layers.}
  \label{fig:comp3}
\end{figure*}

To better understand this result, we visualize the mean of ordered activation value for both original capsnet and the capsnet implemented with our improvement methods(with CI-squash or with PA). To have a clearer interpretation, we plot the largest 1000 activation values(64*12*12 in total) and the remain activation values in two images, shown in Fig.~\ref{fig:comp3}(a) and Fig.~\ref{fig:comp3}(b). From both figures, We can see that both PA and CI-squash are effective at suppressing capsules originally with small activation value, which is consistent with our expectation. We also print the activation map, shown in Fig.~\ref{fig:Comparison_in_activation_map_between_PA_and_nonPA} and Fig.~\ref{fig:Comparison_activation_map_between_CI-squash_and_original_squash}. As from the figures, the capsnet implementing our method has more suppression effect on capsules that are original with small activation value  and can highlight capsules corresponding to object, we can infer and testify our conjecture that by using CI-squash or PA, the capsnet can restrain capsules that are not corresponding to  object and thus encodes relatively irrelevant information. Therefore our methods can improve the mechanism of  activation value assigning to capsules in primary capsule layer, increase the proportion of relevant information following through the network, reduce the conflict between the sensitiveness of capsule network(as will discuss in Section 5.1) and the irrational high activation value distribution existing in primary capsule layer, and finally can lead to a better result.

However, we have experimented with these methods on capsnet with 64 primary capsule layers. The effectiveness of CI-squash and PA may be induced by over redundant capsules. To test the robustness of our methods, we also test them on capsnet with only 8 primary capsule layers, and the results shown in the first column of Tab.~\ref{tab:cifar}. We also generate the mean ordered activation value for this experiment, shown in Fig.~\ref{fig:comp3}(c). From Fig.~\ref{fig:comp3}(c), we can observe the same effect of restraining capsules with small activation value. In addition, with PA(n=6), we can achieve a comparable performance of capsnet with 64 primary capsule layers with only 8 primary capsule layers, which means we may reduce nearly seven-eighths of total parameters and total computations when capsule network is further used in other implementations.

\subsection{5.2 Comparison with common regularization methods}
We have compared PA and CI-squash with weight decay and dropout, results shown in Tab.~\ref{tab:regularization_cifar}. We use CIFAR10 as testing dataset. To save computation resources, for this experiment we use capsnet with 8 primary capsule layers, which has shown to be similarly effective with capsnet with 64 primary capsule layers. For CI-squash, we use a fixed bar value assigned with 6.5. Results shown in Tab.~\ref{tab:regularization_cifar}.

\begin{table}[h]
\centering   
\caption{Comparison between PA, CI-squash, weight decay and dropout (shown as error rate in CIFAR10)}
\begin{tabular}{c|c}
\toprule
Method & error rate(\%) \\
\midrule
capsnet+original squash & 16.15\%  \\
\midrule
capsnet+PA & \textbf{14.31}\% \\
\midrule
capsnet+CI-squash & 14.67\% \\
\midrule
capsnet+weight decay, lamda: 3e-5& 19.15\%\\
\midrule
capsnet+weight decay, lamda: 1e-5& 16.16\%\\
\midrule
capsnet+weight decay, lamda: 3e-6& 16.16\%\\
\midrule
capsnet+dropout, keep dim:0.5 & 17.76\%\\
\midrule
capsnet+dropout, keep dim:0.8 & 17.54\%\\
\midrule
capsnet+dropout, keep dim:0.9 & 16.93\%\\
\bottomrule
\end{tabular}
\label{tab:regularization_cifar}
\end{table}

From Tab.~\ref{tab:regularization_cifar}, we can see that neither of the traditional regularization methods can work on capsule network. As discussed in Section 5.2, CI-squash and PA can be seen as a regularization method for capsule network. Also consider that CI-squash and PA are both easy to implement and the fact that our methods work on various mainstream datasets(also all datasets we have tested the our methods on), therefore we can say that our methods, specifically CI-squash and PA, can be seen as a suitable, simple and efficient regularization method that can be generally used in capsule network. 

\subsection{5.3 Hyperparameters of PA and CI-squash}
We have experimented our method with different hyperparameter, $n$ specifically, of power function. We use the same setting as used on CIFAR10 dataset with 8 primary capsule layers discussed in Section 4.1.  Results are shown in Tab.~\ref{tab:para_of_n_PA}.

\begin{table}[h]
\centering   
\caption{Comparison on different $n$ of PA}
\begin{tabular}{c|c}
\toprule
Method & error rate(\%) \\
\midrule
capsnet+PA($n$=3) &  14.99\%  \\
\midrule
capsnet+PA($n$=6) &  \textbf{14.31}\%\\
\midrule
capsnet+PA($n$=9) &  14.62\%\\
\bottomrule
\end{tabular}
\label{tab:para_of_n_PA}
\end{table}

We only test $n$ with the value of 3, 6 and 9. We can see that when $n$ is assigned with 6 the model can nearly achieve the optimum point. Better performance may be achieved by further accurating to next decimal places. We also test CI-squash with other hyperparameter, like switching to 'Square-Increasing Squash' or 'The Forth Square-Increasing Squash', but CI-squash seems to remain to achieve the regular optimum point.

In addition, as for which method to choose between CI-squash and PA, we suggest that when large primary capsule layers are used(e.g. larger than 32), CI-squash is preferred, while with small primary capsule layers PA is more possible to achieve better performance. 

\section{6. Conclusion}

In this paper, we give an analysis of one main reason why capsule networks cannot achieve compatible performance as conventional CNNs in color image datasets (like CIFAR10), which we call as the conflict between information sensitiveness of capsnet and unreasonable high distribution of activation values in primary capsule layer. Based on this analysis we propose the idea of trying to solve the information sensitiveness problem and introduce our two simple implementation methods of this idea. In addition, we find that the proposed PA and CI-squash can be used as simple, efficient and universal regularizers for capsule networks.
We argue that our two improvement methods are simple and probably not sufficient way to solve the information sensitiveness problem. In the future, we plan to extend the analysis of information sensitiveness and try to explore more efficient way to deal with this problem. 

\printbibliography 
\end{document}